\documentclass[10pt,twocolumn,letterpaper]{article}

\usepackage[pagenumbers]{cvpr} 

\usepackage{pifont}
\newcommand{\xmark}{\ding{55}}

\usepackage{pgfplots}
\usepackage{pgfplotstable}
\usepackage{tikz}

\definecolor{mygreen}{RGB}{93,173,85}

\definecolor{cvprblue}{rgb}{0.21,0.49,0.74}
\usepackage[pagebackref,breaklinks,colorlinks,allcolors=cvprblue]{hyperref}
\usepackage{multirow}
\usepackage{adjustbox}
\def\paperID{2} 
\def\confName{CVPR}
\def\confYear{2025}

\title{Out-of-Distribution Segmentation in Autonomous Driving:\\ Problems and State of the Art }

\author{Youssef Shoeb\\
Continental AG\\
Technische Universität Berlin\\
{\tt\small youssef.shoeb@continental.com}
\and
Azarm Nowzad\\
Continental AG\\
{\tt\small azarm.nowzad@continental.com}
\and
Hanno Gottschalk\\
Technische Universität Berlin\\
{\tt\small gottschalk@math.tu-berlin.de}
}

\begin{document}
\maketitle
\begin{abstract}
In this paper, we review the state of the art in Out-of-Distribution (OoD) segmentation, with a focus on road obstacle detection in automated driving as a real-world application.
We analyse the performance of existing methods on two widely used benchmarks, SegmentMeIfYouCan Obstacle Track and LostAndFound-NoKnown, highlighting their strengths, limitations, and real-world applicability.
Additionally, we discuss key challenges and outline potential research directions to advance the field. Our goal is to provide researchers and practitioners with a comprehensive perspective on the current landscape of OoD segmentation and to foster further advancements toward safer and more reliable autonomous driving systems. 

\end{abstract}    
\section{Introduction}
\label{sec:intro}
The task of semantic segmentation involves assigning each pixel in an image to a predefined class. In recent years, neural networks have become the state-of-the-art approach for semantic segmentation, achieving remarkable accuracy across applications such as autonomous driving, medical imaging, and industrial inspection.
However, a fundamental limitation of these models is their reliance on closed-set assumptions; they are only trained to recognize and classify objects from a predefined set of categories. As a result, when deployed in open-world settings, they may encounter objects or scenes outside this training distribution. In such cases, these models often produce incorrect predictions with high confidence, as they lack an explicit mechanism for identifying unknown inputs~\cite{vaze2022openset}.

This shortcoming is particularly critical in autonomous driving, where safety depends on accurately perceiving all potential hazards on the road. A vehicle must detect and react to unexpected obstacles, such as lost cargo, wildlife, or road debris, even if these objects were never seen during training. To address this challenge, there is a growing need for models that not only segment known objects but also identify and flag unknown ones, a task known as Out-of-Distribution (OoD) segmentation.

Similar to OoD segmentation, other problem formulations, such as \emph{anomaly} and \emph{open-world} segmentation, share similar motivations and methodologies but differ in their definitions. While there is no universally accepted definition of OoD in the literature, we adopt the following distinctions in this survey to guide our discussion. 
Specifically, we define anomalies as rare or unusual instances that occur within a known distribution (e.g., an atypical type of vehicle), while OoD objects stem from entirely unknown semantic categories not presented during training.
While anomalies include instances that may or may not be correctly classified by a model, OoD objects represent a fundamental mismatch between the learned categories of the model and the objects encountered during deployment. 
Meanwhile, open-world segmentation is the task of simultaneously detecting both known and unknown objects during inference.
However, the absence of benchmarks that provide labels for both inlier and unknown classes makes it challenging to distinguish between open-world and OoD segmentation in practice. As a result, it has become common practice to evaluate OoD segmentation performance on an external dataset while assessing inlier performance on the original training dataset, effectively blurring the boundary between OoD and open-world segmentation. Therefore, in this survey, we do not explicitly differentiate between both tasks but acknowledge that, ideally, they should be treated separately, as OoD segmentation methods are typically designed as stand-alone models, whereas open-world segmentation methods are incorporated in a segmentation model. 

In this survey, we provide a holistic overview of recent developments in OoD segmentation, with particular emphasis on road obstacle detection as a practical use case, where identifying novel objects is critical for vehicle safety. We systematically review state-of-the-art methodologies, highlight key limitations, and discuss open research challenges.
Finally, we propose future research directions that we believe will drive the field forward and contribute to safer autonomous systems. 

The remainder of this paper is structured as follows: Section \ref{problem_definition_and_metrics} defines the problem formulation, evaluation metrics, and datasets. Section \ref{sec:sota} reviews state-of-the-art methods, summarising them and briefly mentioning some of their limitations. Section \ref{sec:limits} discusses key limitations of current benchmarks, and Section \ref{sec:discussion} outlines future research directions.

\section{Problem Definition, Metrics \& Dataset }
\label{problem_definition_and_metrics}
\textbf{Problem Definition:} The goal of an OoD segmentation method is to assign an OoD score to each pixel in an image, distinguishing in-distribution from OoD pixels. 
This is achieved by defining a function
\begin{equation}
    s:\mathcal{X} \rightarrow \mathbb{R}^{H\times W}
\end{equation}
where $\mathcal{X}\subseteq [0,1]^{H\times W \times 3}$ represents the space of RGB images with normalized pixel values, and $s(x) \in \mathbb{R}^{H\times W}$ is the OoD score map where each pixel in the input image $x \in \mathcal{X}$ is assigned a real-valued score indicating its likelihood of being OoD. 

\textbf{Evaluation Protocol}: Evaluating OoD segmentation is more complex than standard image-level OoD detection due to the dense, pixel-wise nature of the task. In particular, performance metrics can either be computed per image and then averaged across the dataset or aggregated across all pixels before computing a global metric. In this work, we adopt the latter strategy: all predictions and ground truth labels are collected across the full test set, and metrics are computed on the resulting global distribution of in-distribution and OoD pixels. This approach reduces the influence of image-level variance and reflects the overall effectiveness of the model across the entire dataset.

\textbf{Metrics:} One of the standard metrics for evaluating the separability of binary classification models is the area under the Receiver Operating Characteristic (ROC) curve (AUC-ROC). However, in OoD segmentation, datasets are often highly imbalanced, with significantly fewer OoD pixels compared to in-distribution pixels. In such cases, the area under the precision-recall curve (AUPRC) is preferred as an alternative evaluation metric. Unlike AUC-ROC, which is based on true positive and false positive rates, AUPRC is computed using precision and recall for OoD pixels. This makes it a more informative measure of how well a model identifies OoD pixels without being disproportionately influenced by the large number of in-distribution pixels.
The AUPRC approximates the integral$\int \text{precision}(\delta) \, d\text{recall}(\delta)$ at multiple thresholds. 
For an input image $x \in \mathcal{X}$, let  $\mathcal{Y} \in \{0,1\}^{H\times W}$ denote the binary ground truth mask, where $\mathcal{Y}_{ij} =1$ indicate and OoD pixel. Then the set of ground-truth OoD pixels are defined as $\mathcal{Y}_1 = \{ (i,j) \mid \mathcal{Y}_{i,j} = 1 \}$. 
Similarly, given an OoD score map $s(x)$ and a threshold $\delta \in \mathbb{R}$, the set of predicted OoD pixels is defined as $\hat{\mathcal{Y}}_1(\delta) = \{ (i,j) \mid s(x)_{i,j} > \delta \}$  
The precision and recall at a threshold $\delta$ are then defined as: 
\begin{equation}
    \text{precision}(\delta) = \frac{|\mathcal{Y}_1 \cap \hat{\mathcal{Y}}_1(\delta)|}{|\hat{\mathcal{Y}}_1(\delta)| + \epsilon}, \quad \text{recall}(\delta) = \frac{|\mathcal{Y}_1 \cap \hat{\mathcal{Y}}_1(\delta)|}{|\mathcal{Y}_1| + \epsilon}
\end{equation}
where $\epsilon$ is a small constant to avoid division by zero. 

A more safety-relevant metric is the false positive rate at 95\% true positive rate (FPR$_{95}$). Here, the true positive rate is equal to the recall of the OoD class, and the false positive rate is the number of pixels falsely predicted as OoD over all in-distribution pixels. Formally, FPR$_{95}$ is then calculated as: 
\begin{equation}
    \text{FPR}_{95} = \min_{\delta \in \mathbb{R}} \left\{ \frac{|\hat{\mathcal{Y}}_1(\delta) \cap \mathcal{Y}_0|}{|\mathcal{Y}_0|} \;\Big|\; \text{recall}(\delta) \geq 0.95 \right\}
\end{equation}
where: $ \mathcal{Y}_0 = \{(i,j) \mid \mathcal{Y}_{i,j} = 0\} $ represents the set of in-distribution pixels.

One disadvantage of pixel-level metrics is that they may be biased against small OoD objects, potentially neglecting their contribution to the evaluation. From a practical perspective, one is interested in detecting all OoD objects, regardless of their size \textit{i.e.}, the number of pixels an object covers. Therefore, component-level metrics provide a more holistic evaluation by assessing model performance at the level of components rather than pixels.  

A component is defined as a set of connected pixels that share the same label. 
Let $\mathcal{K} = \{ \mathcal{K}_1,\mathcal{K}_2, \dots, \mathcal{K}_m \}$ denote the set of ground-truth OoD components, where $\mathcal{K}_i$ is the set of connected pixels labelled as OoD in the ground truth mask. Similarly, let $\hat{\mathcal{K}} = \{ \hat{\mathcal{K}}_1,\hat{\mathcal{K}}_2, \dots, \hat{\mathcal{K}}_n \}$ be the set of predicted OoD components. 
The component-wise intersection over union (sIoU) for a ground truth component $k \in \mathcal{K}$ is defined as: 
\begin{equation}
        \text{sIoU}(k) := \frac{|k \cap \hat{\mathcal{K}}(k)|}{|(k \cup \hat{\mathcal{K}}(k)) \setminus \mathcal{A}(k)|} \in [0,1]
\end{equation}
where $\hat{\mathcal{K}}(k) = \bigcup_{\hat{k} \in \hat{\mathcal{K}}, \hat{k} \cap k \neq \emptyset }{\hat{k}}$ and $\mathcal{A}(k) = \bigcup_{k_0 \in \mathcal{K}, k_0 \neq k }{k_0}$. 
The sIoU captures how well a ground-truth component $k$ aligns with predicted components, given a threshold $\tau \in [0,1)$, a target $k \in \mathcal{K}$ is considered a true positive (TP) if sIoU($k$) $> \tau$, and a false negative (FN) otherwise.
A predicted component $\hat{k}$ is considered a False Positive (FP) if its Positive Predictive Value (PPV) is $\leq \tau$, meaning it has insufficient overlap with any ground-truth component. The PPV for $\hat{k} \in \hat{\mathcal{K}}$ is defined as:
\begin{equation}
    \text{PPV}(\hat{k}) := \frac{|\hat{k} \cap \mathcal{K}(\hat{k})|}{|\hat{k}|} \in [0,1] .
\end{equation}

To summarize the TPs, FPs, and FNs, the mean $F_1$-score is averaged across multiple thresholds to summarize the detection performance of a model. The $F_1$-score at a threshold $\tau$ is calculated as follows: 
\begin{equation}
    F_1(\tau) := \frac{2 \cdot \text{TP}(\tau)}{2 \cdot \text{TP}(\tau) + \text{FP}(\tau) + \text{FN}(\tau)} \in [0,1]
\end{equation}

As the numbers of TPs, FPs, and FNs depend on the detection threshold $\tau$, the average F1 score over different $\tau$ is used to give ${\overline{F1}}$ as an evaluation metric that is less affected by the detection threshold.

\begin{table*}[h!]
    \centering
    \adjustbox{max width=\textwidth}{%
    \begin{tabular}{@ {} l l  l l l l l  l l l l @ {}} 
    \toprule
    Method & \multicolumn{5}{c}{Configuration} & \multicolumn{5}{c}{Metrics} \\
      & &  & & & & \multicolumn{2}{c}{Pixel Level}  & \multicolumn{3}{c}{Component Level}\\
    \cmidrule(lr){7-8} \cmidrule(l){9-11} 
           & OE & M2F & GM & UE & Other & $\text{AUPRC}$ $\uparrow$ & $\text{FPR}_{95}$ $\downarrow$ & $sIoU_{gt}$ $\uparrow$  & $PPV$ $\uparrow$ & ${\overline{F1}}$ $\uparrow$ \\
    \midrule
    RbA~\cite{Nayal2023ICCV} & \checkmark & \checkmark & \xmark & \xmark & \xmark & \textbf{95.12} & \textbf{0.08} & 53.34 & 59.08 & 57.44 \\
    
    UEM~\cite{nayal2024likelihoodratiobasedapproachsegmenting} & \checkmark & \xmark & \xmark & \xmark & \checkmark & \underline{94.38} & \underline{0.10} & 57.90 & 41.20 & 43.30 \\

    Mask2Anomaly~\cite{Rai2023ICCV} & \checkmark & \checkmark & \xmark & \xmark & \xmark & 93.22 & 0.20 & 55.72 & 75.42 & 68.15 \\

    UNO~\cite{delic24bmvc} & \checkmark & \checkmark & \xmark & \xmark & \xmark & 93.19 & 0.16 & \textbf{70.97} & 72.17 & \underline{77.65}\\
    EAM~\cite{Grcic23CVPRW} & \checkmark & \checkmark & \xmark & \xmark & \xmark & 92.87 & 0.52 & \underline{65.85} & \underline{76.50} & 75.58 \\
    LR~\cite{shoeb2024segment} & \checkmark & \xmark & \xmark & \xmark & \checkmark & 92.00 & 0.20 & 62.90 & \textbf{81.90} & \textbf{78.40} \\

    UEM (generative)& \checkmark & \xmark & \checkmark & \xmark & \checkmark & 88.30 & 0.40 & 42.60 & 59.20 & 50.90 \\
    
    PixOOD~\cite{Vojir_2024_ECCV} & \xmark & \xmark & \xmark & \xmark & \checkmark & 88.90 & 0.30 & 42.68 & 57.49 & 50.82\\
    RbA & \xmark & \checkmark & \xmark & \xmark & \xmark & 87.85 & 3.33 & 47.44 & 56.16 & 50.42 \\
    
    CLS~\cite{Zhang_AAAI_2024}& \xmark & \checkmark & \xmark & \xmark & \xmark & 87.10 & 0.67 & 44.70 & 53.13 & 51.02 \\

    DenseHybrid~\cite{grcic22eccv} & \checkmark & \xmark & \checkmark & \xmark & \xmark & 87.08 & 0.24 & 45.74 & 50.10 & 50.72 \\
    
    Maximized Entropy~\cite{Chan2021ICCV} & \checkmark & \xmark & \xmark & \checkmark & \xmark & 85.07 & 0.75 & 47.87 & 62.64 & 48.51 \\
    
    DaCUP~\cite{Vojir_2023_WACV} & \xmark & \xmark & \checkmark & \xmark & \xmark & 81.50 & 1.13 & 37.68 & 60.13 & 46.01 \\
    
    ATTA~\cite{gao2023atta} & \checkmark & \xmark & \xmark & \xmark & \checkmark & 76.26 & 2.81 & 43.93 & 37.66 & 36.57 \\
    
    FlowEneDet~\cite{flowenedet} & \xmark & \xmark & \checkmark & \xmark & \xmark & 73.71 & 0.97 & 42.62 & 42.25 & 39.96 \\
    SynBoost ~\cite{di2021pixel} & \xmark & \xmark & \checkmark & \xmark & \xmark & 71.34 & 3.15 & 44.28 & 41.75 & 37.57 \\
    
    DOoD~\cite{GalessoECCV2024} & \xmark & \xmark & \checkmark & \xmark & \xmark & 64.30 & 2.60 & 30.30 & 33.10 & 24.10 \\
    
    Image Resynthesis~\cite{lis2019detecting} & \xmark & \xmark & \checkmark & \xmark & \xmark & 37.71 & 4.70 & 16.61 & 20.48 & 8.38 \\
    
    JSRNet~\cite{vojir2021road}  & \xmark & \xmark & \checkmark & \xmark & \xmark & 28.09 & 28.86 & 18.55 & 24.46 & 11.02 \\ 
    
    PGN~\cite{maag2023pixel} & \xmark & \xmark & \xmark & \checkmark & \xmark & 16.52 & 19.69 & 19.42 & 14.89 & 7.39 \\
    
    Maximum Softmax~\cite{hendrycks2017a} & \xmark & \xmark & \xmark & \checkmark & \xmark & 15.72 & 16.60 & 19.72 & 15.93 & 6.25 \\
    
    PEBAL~\cite{tian2022pixel} & \checkmark & \xmark & \checkmark & \xmark & \xmark & 4.98 & 12.68 & 29.91 & 7.55 & 5.54 \\
    
    MC Dropout~\cite{mukhoti2018evaluating} & \xmark & \xmark & \xmark & \checkmark & \xmark & 4.88 & 50.31 & 5.49 & 5.77 & 1.05 \\
    
    Ensemble~\cite{lakshminarayanan2017simple} & \xmark & \xmark & \xmark & \checkmark & \xmark & 1.06 & 77.20 & 8.63 & 4.71 & 1.28 \\
    
    Embedding Density~\cite{blum2019fishyscapes} & \xmark & \xmark & \checkmark & \xmark & \xmark & 0.82 & 46.38 & 35.64 & 2.87 & 2.31  \\
    
     \bottomrule
    \end{tabular}}
    \caption{\textbf{Quantitative Results on SMIYC-OT}: Comparison of current methods on the SMIYC-OT dataset. Methods are sorted from top to bottom according to their AP. The best result for each metric is highlighted in \textbf{bold}, and the second best is \underline{underlined}. Methods are categorized into four groups: Mask2Former-based (M2F), Uncertainty Estimation (UE), Generative Models (GM), and others.
 }
    \label{tab:main_table}
\end{table*}

\textbf{Dataset:} We evaluate the performance of various methods on the SegmentMeIfYouCan Obstacle Track (SMIYC-OT) and LostAndFound-NoKnown (L\&F) datasets~\cite{chan2021segmentmeifyoucan}. Both benchmarks assess the ability to segment OoD objects on the road that do not belong to any of the 19 predefined classes in the Cityscapes dataset~\cite{cordts2016cityscapes}.
The SMIYC-OT dataset comprises 327 test images containing a total of 388 OoD instances from 31 distinct obstacles captured under diverse real-world conditions, including clear weather, night scenes, and snowy environments. Notably, SMIYC-OT introduces a significant domain shift compared to Cityscapes. The LostAndFound-NoKnown benchmark is a filtered and refined version of the LostAndFound dataset~\cite{pinggera2016lost}, designed to better align with the OoD segmentation task. It consists of 1,043 test images with a total of 1,709 OoD instances from seven categories. Unlike SMIYC-OT, L\&F shares the same domain as Cityscapes in terms of lighting, camera settings, and geographic location. Both datasets provide pixel-level annotations for the road and obstacles in the vehicle’s path.

While other OoD segmentation datasets exist, such as Fishyscapes~\cite{blum2021fishyscapes} and RoadAnomaly~\cite{lis2019detecting}, we chose not to include them in this survey for specific reasons. Fishyscapes Static includes synthetic anomalies overlaid from unrelated datasets, which resemble the process of outlier exposure and may lead to models overfitting to the synthetic nature of these objects rather than generalizing to real-world unknowns. RoadAnomaly, on the other hand, introduces a considerable domain shift from Cityscapes due to differences in camera viewpoint and height relative to the road surface, which can result in unrealistic scenarios for standard autonomous driving applications. Consequently, we focus on SMIYC-OT and L\&F as they provide a more balanced and representative evaluation setting for the real-world OoD segmentation task.

\section{Current State of the Art}
\label{sec:sota}

\begin{table*}[ht]
    \centering
    \adjustbox{max width=\textwidth}{%
    \begin{tabular}{@ {} l l  l l l l l  l l l l @ {}} 
    \toprule
    Method & \multicolumn{5}{c}{Configuration} & \multicolumn{5}{c}{Metrics} \\
      & &  & & & & \multicolumn{2}{c}{Pixel Level}  & \multicolumn{3}{c}{Component Level}\\
    \cmidrule(lr){7-8} \cmidrule(l){9-11} 
           & OE & M2F & GM & UE & Other & AUPRC $\uparrow$ & $FPR$ $\downarrow$ & $sIoU_{gt}$ $\uparrow$  & $PPV$ $\uparrow$ & ${\overline{F1}}$ $\uparrow$ \\
    \midrule
    
    PixOOD~\cite{Vojir_2024_ECCV} & \xmark & \xmark & \xmark & \xmark & \checkmark & \textbf{85.07} & 4.46 & 30.18 & \underline{78.47} & 44.41\\
    
    LR~\cite{shoeb2024segment} & \checkmark & \xmark & \xmark & \xmark & \checkmark & \underline{83.70} & 100.00 & \textbf{49.70} & \textbf{95.90} & \textbf{72.60} \\

    DaCUP~\cite{Vojir_2023_WACV} & \xmark & \xmark & \checkmark & \xmark & \xmark & 81.37 & 7.36 & 38.34 & 67.29 & 51.14 \\

    UEM~\cite{nayal2024likelihoodratiobasedapproachsegmenting} & \checkmark & \xmark & \xmark & \xmark & \checkmark & 81.04 & \textbf{1.45} & 39.49 & 55.77 & 46.54 \\

    FlowEneDet~\cite{flowenedet} & \xmark & \xmark & \checkmark & \xmark & \xmark & 79.75 & 2.92 & 43.79 & 52.83 & 48.05 \\

    DOoD~\cite{GalessoECCV2024} & \xmark & \xmark & \checkmark & \xmark & \xmark & 79.50 & 3.70 & 27.70 & 40.60 & 30.5 \\  
    
    RbA~\cite{Nayal2023ICCV} & \checkmark & \checkmark & \xmark & \xmark & \xmark & 79.25 & 22.57 & 45.56 & 68.90 & \underline{59.31} \\

    DenseHybrid~\cite{grcic22eccv} & \checkmark & \xmark & \checkmark & \xmark & \xmark & 78.67 & \underline{2.12} & \underline{46.90} & 52.14 & 52.33 \\
    
    Maximized Entropy~\cite{Chan2021ICCV} & \checkmark & \xmark & \xmark & \checkmark & \xmark & 77.90 & 9.70 & 45.90 & 63.06 & 49.92 \\
    
    JSRNet~\cite{vojir2021road}  & \xmark & \xmark & \checkmark & \xmark & \xmark & 74.17 & 6.59 & 34.28 & 45.89 & 35.97 \\ 
   
    PGN~\cite{maag2023pixel} & \xmark & \xmark & \xmark & \checkmark & \xmark & 69.30 & 9.80 & 50.00 & 44.80 & 45.40 \\ 
    
    Embedding Density~\cite{blum2019fishyscapes} & \xmark & \xmark & \checkmark & \xmark & \xmark & 61.70 & 10.36 & 37.75 & 35.21 & 27.55  \\

    Image Resynthesis~\cite{lis2019detecting} & \xmark & \xmark & \checkmark & \xmark & \xmark & 57.08 & 8.82 & 27.16 & 30.69 & 19.17 \\

    MC Dropout~\cite{mukhoti2018evaluating} & \xmark & \xmark & \xmark & \checkmark & \xmark & 36.78 & 35.55 & 17.35 & 34.71 & 12.99 \\

    Maximum Softmax~\cite{hendrycks2017a} & \xmark & \xmark & \xmark & \checkmark & \xmark & 30.14 & 33.20 & 14.20 & 62.23 & 10.32 \\

    Ensemble~\cite{lakshminarayanan2017simple} & \xmark & \xmark & \xmark & \checkmark & \xmark & 2.89 & 82.03 & 6.66 & 7.64 & 2.68 \\
    
     \bottomrule
    \end{tabular}}
    \caption{\textbf{Quantitative Results on L\&F}: Comparison of current methods on the L\&F dataset. Methods are sorted from top to bottom according to their AP. The best result for each metric is highlighted in \textbf{bold}, and the second best is \underline{underlined}. Methods are categorized into four groups: Mask2Former-based (M2F), Uncertainty Estimation (UE), Generative Models (GM), and others.
 }
    \label{tab:lf_table}
\end{table*}

Table \ref{tab:main_table} and \ref{tab:lf_table} summarise the performance of different methods on the SMIYC-OT and LostAndFound-NoKnown benchmarks, respectively. We report both pixel-level and component-level metrics for each method. For comparing the different methods, we categorize the methods into four distinct groups:
\begin{enumerate}
    \item Mask2Former-based (M2F): Methods that leverage special features from the Mask2former architecture \cite{Cheng2022CVPR}.

    \item Uncertainty Estimation (UE): Approaches that rely on confidence or uncertainty estimates. 
    
    \item Generative Models (GM): Methods that utilize generative modelling for OoD scoring.
    
    \item Other: Methods that do not fall into the above categories but offer unique approaches to OoD detection.
\end{enumerate}
In addition, we also denote if the model used any outlier data to improve OoD detection performance, a common technique referred to as \emph{outlier exposure} (OE) ~\cite{hendrycksdeep}. In the following section, we start by discussing outlier exposure, followed by the methodology behind each group of methods and some of their limitations.

\subsection{Outlier Exposure}
In the setting where one has knowledge about potential OoD objects a network may encounter, one can improve separation between in-distribution and OoD objects using proxy OoD objects. Proxy OoD data, often referred to as \emph{known unknowns}, are samples that do not belong to the in-distribution classes but resemble potential OoD objects the model might encounter in deployment. Proxy OoD data can then be used to regularize the feature space of the model by learning representations for unknown objects; this technique is referred to as \emph{outlier exposure}. The assumption here is that exposure to proxy OoD during training can help the model generalize to unseen OoD objects in deployment.

For OoD segmentation, outlier exposure has been applied in various ways to enhance OoD detection performance. A common approach involves using a subset of objects from COCO~\cite{lin2014microsoft} or ADE20K~\cite{zhou2017scene} as proxy OoD objects. 
These objects are then pasted into in-distribution images during a fine-tuning stage, either as cutouts pasted on the inlier data or as complete images.
Maximized Entropy~\cite{Chan2021ICCV} encourages higher entropy predictions on proxy OoD samples to prevent overconfidence in uncertain regions, while RbA~\cite{Nayal2023ICCV} and PEBAL~\cite{tian2022pixel} explicitly train the model to produce lower logit scores for these proxy OoD samples.
The general idea is to train the model to learn certain heuristics that can later be used to detect out-of-distribution inputs. Other approaches explicitly train the model on the proxy OoD samples to model unknown objects as a separate class.

While proxy outliers in road obstacle detection can be justified by defining OoD objects as entities not present in the training set, potentially capturing a broad notion of ``objectness", this strategy raises notable concerns. The primary issue is the risk of overfitting to seen examples, which may limit the model’s ability to generalize to truly novel obstacles in real-world scenarios. Moreover, the variability in datasets used for outlier exposure across different methods calls into question how the choice of outliers impacts the effectiveness of these approaches.
Despite its importance, little research explores how outlier selection impacts OoD detection, particularly given the limited object categories in current benchmarks.
We highlight the need for further exploration to determine optimal strategies for selecting proxy data for outlier exposure.

\begin{table*}[h]
    \centering
    \resizebox{\linewidth}{!}{
    \adjustbox{max width=\textwidth}{%
    \begin{tabular}{l l l l l l l l l l l} 
    \toprule
    & \multicolumn{5}{c}{SMIYC-OT}  & \multicolumn{5}{c}{LostAndFound-NoKnown}\\
    \cmidrule(l){2-6} \cmidrule(l){7-11}
     & \multicolumn{2}{c}{Pixel Level}  & \multicolumn{3}{c}{Component Level} & \multicolumn{2}{c}{Pixel Level}  & \multicolumn{3}{c}{Component Level}  \\
    \cmidrule(lr){2-3} \cmidrule(l){4-6} \cmidrule(lr){7-8} \cmidrule(l){9-11} 
           Method & AUPRC$\uparrow$ & $FPR$ $\downarrow$ & $sIoU_{gt}$ $\uparrow$  & $PPV$ $\uparrow$ & ${\overline{F1}}$ $\uparrow$ & AUPRC$\uparrow$ & $FPR$ $\downarrow$ & $sIoU_{gt}$ $\uparrow$  & $PPV$ $\uparrow$ & ${\overline{F1}}$ $\uparrow$ \\
    \midrule
     
     MaxLogit & 42.73 & 4.26 & 28.28 & 33.97 & 21.89 & 60.19 & 12.88 & 50.63 & 38.98 & 41.32 \\
    
     Softmax Entropy & 42.73 & 4.27 & 24.26 & 25.35 & 15.04 & 36.87 & 14.86&  35.71&  43.92 &  30.37\\

    \hline
     Maximized Entropy  & 81.51 & 0.99 & 46.11 & 36.16 & 38.92 & 85.42 & 1.15 & 57.41 & 53.84 & 58.80 \\
    
     \bottomrule
    \end{tabular}}}
    \caption{ Performance comparison of simple OoD detection methods on a DINOv2 encoder with a feature pyramid network decoder (83~\% mIoU on Cityscapes). Methods below the line used proxy OoD data from MS-COCO for outlier supervision.  
 }
    \label{tab:ue_table}
\end{table*}

\subsection{Uncertainty Estimation}
A neural network with a softmax output layer can be interpreted as a statistical model $f_{\theta}(y|x)$, which assigns a probability distribution over the $n$ class labels $y\in \mathcal{C}=\{y_1,\dots,y_n\}$ for each pixel in the image. This distribution depends on the parameters $\theta$ and the input $x$. Some of the first approaches proposed for OoD segmentation were to use uncertainty metrics on these probability scores like: max probability~\cite{hendrycks2017a}, softmax-entropy~\cite{Chan2021ICCV}, approximations over the posterior distribution of the weights of the model~\cite{mukhoti2018evaluating} or predictive entropy of deep ensembles~\cite{lakshminarayanan2017simple} to estimate the OoD score of pixels.

While these uncertainty-based metrics capture different aspects of model confidence, it is important to distinguish between epistemic, aleatoric, and predictive uncertainty.  
This decomposition is not universally agreed upon across all disciplines but follows a widely accepted framework in the probabilistic modelling community, as introduced in~\cite{gal2016uncertainty}.
In this framework, epistemic uncertainty, also known as model uncertainty, arises from the lack of knowledge about the true model parameters and can be reduced with more training data; this is typically the uncertainty one is interested in for OoD detection, as it can be reduced by collecting more data. The epistemic uncertainty is quantified as the expected information gain between the network parameters $\theta$ and the output $\mathbb{I}[Y;\theta|x,\mathcal{D}]$. On the other hand, aleatoric uncertainty represents inherent noise in the data, such as sensor noise, and cannot be reduced by collecting more data.
Aleatoric uncertainty $\mathbb{E}_{p(\theta|\mathcal{D})}[\mathbb{H}[Y|x,\theta]]$, represents the expected entropy of the output distribution given fixed model parameters. Predictive uncertainty is the total uncertainty in the model’s predictions and is composed of both epistemic and aleatoric uncertainty: 
\begin{equation}
    \mathbb{H}[Y|x,\mathcal{D}] = \mathbb{I}[Y;\theta|x,\mathcal{D}] + \mathbb{E}_{p(\theta|\mathcal{D})}[\mathbb{H}[Y|x,\theta]].
    \label{eq:predictive_decomposition}
\end{equation}

Eq.\ref{eq:predictive_decomposition} highlights a key limitation of using predictive entropy for OoD detection: it does not distinguish between epistemic and aleatoric uncertainty. As a result, high predictive entropy does not necessarily indicate an OoD sample; it may also correspond to an ambiguous in-distribution input. Thus, predictive entropy is only a reliable OoD measure in the absence of ambiguous samples. While deep ensembles and Bayesian models improve epistemic uncertainty estimation, predictive entropy alone remains an unreliable OoD detection metric since it conflates epistemic and aleatoric uncertainty. For a single deterministic model, methods like softmax entropy can be used as a proxy for uncertainty. However, empirical findings from~\cite{kirsch2021pitfalls} suggest that while deep ensembles can effectively capture epistemic uncertainty via mutual information, softmax entropy in a deterministic model fails to do so. In practice, separating the two components of uncertainty is often very difficult, as they are strongly correlated in many practical settings~\cite{mucsanyi2025benchmarking}.

Training with outlier exposure reduces epistemic uncertainty by incorporating OoD examples into the learning process, effectively shifting uncertainty from epistemic to aleatoric. As a result, softmax and predictive entropy become more reliable uncertainty measures. However, this raises a fundamental question: if a model has been trained with outliers, are these samples still truly OoD? Furthermore, could excessive exposure to outliers lead to overgeneralization, reducing the ability of the model to separate inliers from outliers?

An alternative approach for quantifying pixel-level uncertainty in OoD segmentation was proposed in PGN~\cite{maag2023pixel}. Their approach computes gradient-based uncertainty scores at the pixel level by leveraging loss gradients during inference. The intuition behind this approach is that OoD pixels exhibit high gradient magnitudes, as they do not align well with any in-distribution class.

Uncertainty-based methods generally rank at the bottom of the benchmark (with the exception of Maximized Entropy~\cite{Chan2021ICCV} which utilized outlier supervision and a post-processing method called \emph{Meta-Classification}~\cite{rottmann2020prediction} to eliminate false-positives), this is because closed-set models are not necessarily calibrated, often leading to overconfident predictions on unseen classes~\cite{nguyen2015deep}.
In Table~\ref{tab:ue_table}, we evaluate simple uncertainty estimation methods on a DINOv2~\cite{Oquab2024TMLR} encoder with a FPN decoder~\cite{lin2017feature}, which achieves an 83~\% mIoU on Cityscapes. We find that unnormalized uncertainty measures, such as MaxLogit~\cite{vaze2022openset}, significantly outperform maximum softmax. Additionally, we quantify the benefits of outlier exposure and demonstrate its substantial impact on performance. However, outlier exposure led to a $\approx$ 2~\% mIoU drop in inlier performance. Note that meta-classification is not included in these experiments.

\subsection{Generative Models}
An intuitive approach for OoD detection that does not require any labelled samples is to fit a density model to the in-distribution samples and then use the inverse of the likelihood $p(x)$ as an OoD score. One of the first approaches for OoD segmentation~\cite{blum2019fishyscapes} followed this approach and used a normalizing flow~\cite{dinh2017density} to approximate the density of intermediate embeddings of a DeepLabv3+ segmentation model~\cite{chen2018encoder}. This approach was later extended in FlowEneDet~\cite{flowenedet}, where the authors use a 2D Glow-like~\cite{kingma2018glow} architecture trained using correctly classified pixels as positive samples and the misclassified pixels as negative samples. The proposed architecture in FlowEneDet is an extra network that can be added to any semantic segmentation network. Beyond pure generative methods, an alternative perspective views discriminative models with softmax classifiers as energy-based models for the joint distribution $p(x,y)$ through the LogSumExp scoring function on the logits~\cite{grathwohlyour}. This interpretation led to approaches like PEBAL~\cite{tian2022pixel}, which improved upon simple maximum-logit scoring for OoD segmentation. More recently, hybrid approaches such as DenseHybrid~\cite{grcic22eccv} have combined generative modelling of the inlier data with discriminative training of negative samples via outlier exposure to enhance OoD segmentation. 

The main disadvantage of using generative models for OoD detection via density estimation is that they do not explicitly differentiate between outliers within the in-distribution data and true OoD samples. Since the in-distribution data itself may contain outlier points (e.g., samples from the tail of the distribution), the model might assign low likelihoods to these points, potentially misclassifying them as OoD. Additionally, it has been empirically shown that generative models trained on one dataset assign a higher likelihood to samples from OoD datasets~\cite{Nalisnick2019ICLR, Ren2019NeurIPS, kirichenko2020normalizing, zhang2021understanding} in image classification tasks.  
The authors in~\cite{zhangfalsehoods} argue that the principled way of doing OoD detection is using the likelihood ratios between a proxy OoD estimate and the in-distribution estimate. In one of their formulations, the authors of~\cite{nayal2024likelihoodratiobasedapproachsegmenting} use the likelihood ratios with the inlier and OoD estimates obtained using a variant of GMMseg~\cite{liang2022gmmseg}. This method is currently the highest-performing generative model on the benchmark. However, the authors found that using a discriminative model in the likelihood ratios resulted in better performance.

Another approach for using generative models for OoD segmentation involves using generative models like variational autoencoders~\cite{kingma2014stochastic} or conditional generative adversarial networks~\cite{8579015} for re-synthesising an image from its predicted semantic map and comparing it with the original image~\cite{lis2019detecting}. The discrepancies between input and generated outputs are then used as an OoD score.
This approach was further expanded to integrated uncertainty estimation with image re-synthesizing to refine the OoD segmentation in SynBoost~\cite{di2021pixel}.
Building on the idea of detecting OoD objects based on the difference between input and generated output, JSRNeT~\cite{vojir2021road} proposes a generative discriminative reconstruction module that can be added on top of a fixed semantic segmentation network. However, instead of generating a complete image, this reconstruction module learns the appearance of the road. This method was then extended in DaCUP~\cite{Vojir_2023_WACV}, which introduced an embedding bottleneck to better model multi-modal road appearances and improve training data utilization. Additionally, a distance-based embedding scoring feature was introduced to refine anomaly detection, along with an inpainting module that filters out false positives by identifying regions reconstructable from their surroundings. These improvements were integrated into a coupling module that combines multiple anomaly estimation signals to enhance overall performance. Diffusion models have also been explored for OoD segmentation in DOoD~\cite{GalessoECCV2024}, where the denoising score is used to compute per-pixel OoD scores. 

A key disadvantage of image re-synthesis and reconstruction approaches is their high inference time. These methods require multiple processing steps: predicting a semantic map, generating a synthesized or reconstructed image, and then computing discrepancies between the input and the output. This additional computational overhead makes real-time applications particularly challenging.

\subsection{Mask2Former}
Five of the top seven methods on this benchmark depend on the Mask2former~\cite{Cheng2022CVPR} architecture for detecting OoD objects. 
Mask2former is a mask-level segmentation architecture that decouples object localization and classification by splitting the task into two steps.
Given an $H \times W$ sized image, Mask2former computes $N$ pairs $\{ (\mathbf{m}_i, \mathbf{p}_i) \}_{i=1}^N$, where $\mathbf{m}_i \in [0,1]^{H \times W}$ are mask predictions associated with some semantically related regions in the input image and $\mathbf{p}_i \in [0,1]^{K+1}$ class probabilities classifying to which semantic category the mask $\mathbf{m}_i$ belongs to.
Here, the masks can be assigned to one of the $K$ known Cityscapes classes or to one auxiliary void class.
The final semantic segmentation inference is carried out by an ensemble-like approach over the pairs $\{ (\mathbf{m}_i, \mathbf{p}_i) \}_{i=1}^N$ yielding pixel-wise class scores 
\begin{equation}
    \mathbf{q}[h,w,k] = \sum_{i=1}^N \mathbf{p}_i(k) \cdot \mathbf{m}_i[h,w] ~~\in [0,N]
\end{equation}
for image pixel locations $h = 1,\ldots,H, w=1,\ldots,W$ and classes $k=1,\ldots,K$. The authors of RbA~\cite{Nayal2023ICCV} noticed that each object query acts as a one-vs-all classifier and that OoD objects can be detected by determining whether a pixel is associated to any of the $k$ known classes. The OoD score is defined as: \begin{equation}\label{eq:anomaly_score_rba}
    \mathbf{RbA}[h,w] = -\sum_{k=1}^K \phi ( \mathbf{q}[h,w,k] ) ~~\in [0,K]
\end{equation}
with $\phi$ being the $\tanh$ activation function.

Other methods that utilize the mask-level architecture of Mask2former and the attention-based transformer decoder include Mask2Anomaly~\cite{Rai2023ICCV}, EAM~\cite{Grcic23CVPRW}, and UNO~\cite{delic24bmvc}. Mask2Anomaly uses a contrastive loss in the outlier supervision process to separate known masks from proxy OoD masks. EAM ensembles region-wide outlier scores from the mask and pixel classifiers in the network architecture. To boost performance, they utilize the additional void class included in the transformer decoder in the outlier supervision process to instruct all masks to avoid the proxy OoD examples.  
UNO extends on EAM but uses an additional object query in the transformer decoder to learn the OoD objects. The final OoD score is an ensemble of the void and proxy OoD object scores.

In CSL~\cite{Zhang_AAAI_2024}, the authors introduce a class-agnostic segmentation framework that enhances OoD detection by integrating structural constraints into existing methods. CSL extends an adjusted version of Mask2Former with a base teacher network and an MLP, generating per-pixel distributions that are fused with inlier region proposals to produce the final OoD prediction. The framework employs soft assignment and mask-split preprocessing to refine segmentation boundaries and improve generalization to unseen classes.

One of the main disadvantages of these approaches is that they are specific to the Mask2former architecture and the transformer decoder, which acts as a one-vs-all classifier. Therefore, these methods are usually not transferable to other architectures. Additionally, RbA, UNO, and EAM utilized the Swin-L~\cite{liu2021swin} backbone, which has $\approx 200$M parameter, making them impractical for real-time deployment, particularly in resource-constrained environments like autonomous vehicles. 

\subsection{Other}
One of the main limitations of the methods that utilize outlier supervision is that the fine-tuning step decreases the performance of the original inlier model. In our own experiments, maximized entropy decreased the inlier performance by 1.6~\% mIoU points, and RbA decreased by 2.5~\% mIoU. In many practical scenarios, this is highly undesirable as OoD objects typically occur far less often than inlier objects. Instead of adjusting the inlier parameters of the model during the outlier supervision process, the authors of UEM~\cite{nayal2024likelihoodratiobasedapproachsegmenting} proposed an external module explicitly designed for OoD segmentation; this allows to keep the original inlier performance intact while performing very well in detecting OoD objects. The authors also suggested using the Likelihood Ratios between the learned distribution of outliers from the proxy OoD dataset and that of the original inlier model to calculate the OoD score. The authors found that using the feature representation of self-supervised pre-trained models like DinoV2~\cite{Oquab2024TMLR} resulted in better modelling of OoD objects than fully supervised models like Swin~\cite{Liu2021ICCV} or contrastive image-language pre-trained models like CLIP~\cite{Radford2021ICML}

Motivated by the use of likelihood ratios in UEM, the authors of LR~\cite{shoeb2024segment} extend the idea and use features from the foundation model Segment Anything Model~\cite{Kirillov_2023_ICCV} (SAM) for OoD segmentation. One of the main advantages of SAM is that it moves away from pixel-level reasoning and reasons about segment-level masks, which results in more semantically meaningful segmentations. The main limitation of this approach is that SAM is limited to prompted prediction, and if an object is not prompted, it will not be assigned a mask. Other approaches~\cite{nekrasov2023ugains} have also proposed using SAM but prompt the predictions based on an existing OoD segmentation model to refine the pixel predictions; this removes some of the false positives and merges some of the predictions to get more coherent segments. However, if the initial OoD segmentation network misses an object, SAM will not detect it. 

In PixOOD~\cite{Vojir_2024_ECCV}, the authors propose an OoD segmentation method that does not require any outlier exposure. Their approach models the intra-class in-distribution variability through an online data condensation algorithm and consists of three components: 1) extracting the patch feature representation from a DinoV2 model, 2) building a 2D-projection space for each in-distribution class, and 3) finding calibrated in-distribution/OoD decision strategy. The framework builds on the methodology introduced in~\cite{Vojir_2023_ICCV} but extends the approach for pixel-level predictions. A similar direction is explored in~\cite{GAB23}, where the authors apply a k-nearest-neighbour-based strategy based on patch feature representations for dense OoD detection.

In ATTA~\cite{gao2023atta}, the authors address the challenge of OoD segmentation under domain shift by proposing a dual-level OoD detection framework. Their method simultaneously handles domain and semantic shifts to improve OoD segmentation robustness. ATTA detects domain shifts at the first level by leveraging global low-level features, enabling selective adaptation of the model to unseen domains. At the second level, it identifies OoD pixels caused by semantic shifts using dense high-level feature maps. The framework employs a ($\mathcal{C}+1$)-class probabilistic classifier, where the first $\mathcal{C}$ classes represent the in-distribution categories, and an additional class is designated for OoD objects. ATTA utilizes an anomaly-aware self-training procedure to refine OoD predictions dynamically. This post hoc adaptation can be applied to various OoD segmentation models, though it does introduce additional computational overhead, nearly doubling inference time.

\section{Limitations}
\label{sec:limits}
While the benchmarks provide a structured framework for assessing the OoD segmentation capabilities of different methods, several design choices and constraints raise concerns about their suitability for real-world deployment. In this section, we analyze some of the benchmark's key limitations.

\textbf{Region of Interest:} Both benchmarks limit the evaluation of detection OoD objects to a predefined region of interest (i.e., the road). This design choice ensures that only the objects of interest are evaluated. In practice, this predefined region is often obtained from an auxiliary network that segments the drivable area, effectively reducing the OoD segmentation task to a foreground-background segmentation problem~\cite{you2025focus}. Some methods explicitly utilize this assumption~\cite{marschall2024multi,vojir2021road, Vojir_2023_WACV, shoeb2024segment}, while others did not make direct use of this predefined region of interest assumption. For road obstacle detection, an open question remains: is it more efficient to use a segmentation network with inherent OoD detection capabilities or to first generate a region of interest and then apply a foreground-background segmentation approach to detect OoD objects? 

\textbf{On Small Objects:} Another limitation of the benchmark design is the removal of small predicted regions below a predefined threshold of  50 pixels. While this filtering step may reduce noise and stabilize evaluation metrics, it undermines one of the key advantages of segmentation-based methods—their ability to detect small anomalous objects. OoD objects in autonomous driving scenarios, such as lost cargo or road debris, are often safety-critical, and segmentation models are specifically designed to capture fine-grained spatial details. By discarding small predictions, the benchmark may underestimate the effectiveness of methods capable of detecting such fine structures. Moreover, this design choice does not adequately penalize models that produce fragmented false positive predictions, as small, scattered errors may be ignored rather than contributing to the overall error metric. 

\textbf{Threshold Selection:} One limitation of existing benchmarks for OoD segmentation is their reliance on threshold-free metrics (e.g., AUPRC), which, while useful for assessing performance independent of a specific threshold, are not directly applicable to downstream tasks. In practice, a threshold must be selected to make reliable decisions, and ideally, it should effectively separate in-distribution and out-of-distribution regions with high confidence.
Furthermore, the SMIYC-OT benchmark assesses component-level performance for practical reasons outlined in Section \ref{problem_definition_and_metrics}. However, their evaluation derives an optimal threshold from pixel-level metrics using the optimal pixel-level F1 score to assess component-level performance. This approach is troublesome for deployment scenarios, as ground-truth labels for OoD objects are often not available in large numbers for calibrating the OoD score, making it difficult to select an optimal threshold in real-world settings. Moreover, this methodology obscures a critical limitation of some approaches: a narrow range of effective OoD scores. These methods are highly sensitive to threshold selection, which means small variations can significantly impact their performance. This sensitivity is not adequately captured with current metrics, which may misrepresent the robustness of different approaches. Metrics like the Maximal Detection Margin~\cite{ackermann2023maskomaly}, are a step in the right direction to mitigate some of these issues. 

\section{Future Directions}
\label{sec:discussion}
Although OoD segmentation on benchmark data sets has reached a certain level of maturity and offers a broad choice of performant models, the question arises of how to proceed with the research on OoD segmentation. Here, we provide a non-exhaustive list of topics. 

\textbf{Downstream Tasks:} The use case of OoD Segmentation should be sharpened. There seems to be little research on how OoD segmentation influences downstream tasks like building environmental models or planning in automated driving or robotics~\cite{wormann2022knowledge}. Intuitively, it is clear that OoD objects should not be hit. But as there is little work on the suspected behaviour of an OoD object, how to avoid something we don't know might be complicated. Avoiding an animal on the road requires a different response than avoiding lost cargo.
Beyond safety concerns in autonomous systems, OoD segmentation can also play a crucial role in self-monitoring AI-based perception algorithms and retrieval of unusual objects \cite{shoeb2024have}. This can lead to a better understanding of the operational domain of deployment (ODD). Research on how to build continuous learning strategies on top of the retrieved data is still in its infancy; see, however, see~\cite{uhlemeyer2022towards,uhlemeyer2023detecting, shoeb2025adaptiveneuralnetworksintelligent} for first steps. 

In addition to detection, context-aware decision-making is essential: not all OoD objects pose the same level of risk. Thus, it becomes crucial to interpret the OoD detection in light of the surrounding scene context and possible future interactions. This requires integrating scene understanding and behaviour prediction with the OoD segmentation pipeline. For instance, distinguishing between a static roadside billboard and a moving pedestrian in an unknown outfit directly influences the vehicle’s response. Such scene-level understanding supports more adaptive and risk-sensitive planning strategies in real-world applications.

\textbf{Real-time Deployment:} Recently, academic OoD segmentation research has profited much from large-scale foundation models. However, OoD segmentation in the wild requires the deployment of OoD segmentation algorithms on edge devices.
To bridge the gap between academic research and real-world deployment, future work must address key efficiency challenges: reducing inference times, ensuring compatibility with multi-task networks, and optimizing OoD models for edge devices through quantization and knowledge distillation. A balance between computational efficiency and segmentation accuracy is critical for industrial adoption

\textbf{New Datasets:} As presently available benchmark datasets have saturated, new datasets should emerge and lead the field into the future. 
The present state of OoD detection suffers from under-complex street scenes \cite{bogdoll2023perception}. This is mainly due to safety requirements in recording OoD data sets. In this way, the effects of immersion of OoD objects into ID scenes, occlusion, and dynamical OoD objects are still to be integrated into contemporary OoD research.
The next frontier in OoD segmentation research requires datasets that reflect real-world complexities incorporating temporal dynamics, multimodal sensor inputs with a camera, LiDAR and RaDAR at least, but potentially also with depth cameras, multi-spectral imaging and acoustic sensors.
It would be desirable if semantic labels were available to measure domain performance. One way to achieve this at a limited cost would be to utilize pseudo labels from highly performant in domain networks \cite{hummer2024strong}. However, this induces a bias towards utilizing the same network, and this might stand in the way of future improvement of both ID and OoD segmentation.     

As we documented in this survey, the rapid progress in OoD segmentation, driven by benchmark datasets, has led to an impressive state-of-the-art. However, the true potential of this field lies in its real-world deployment: ensuring safety in autonomous systems, improving AI-driven decision-making, and pushing the boundaries of open-world perception. Looking into the future, integrating multimodal learning, continuous adaptation, and real-time deployment will be key to unlocking the next generation of OoD segmentation models used in automated driving systems.

\textbf{Acknowledgement:} The research leading to these results is funded by the German Federal Ministry for Economic Affairs and Climate Action within the project ``just better DATA''.

{
    \small
    \bibliographystyle{ieeenat_fullname}
    \bibliography{main}
}


\end{document}